\documentclass[runningheads]{llncs}
\usepackage[T1]{fontenc}
\usepackage[font=small]{caption} 
\usepackage{url}
\usepackage{graphicx}
\usepackage{amsmath, amssymb, bm}
\usepackage{booktabs}
\usepackage{tikz}
\usetikzlibrary{positioning}
\usepackage{caption}
\usepackage{subcaption}
\usepackage{multirow}
\usepackage{makecell}
\usepackage{algorithm}
\usepackage{algorithmic}
\usepackage{hyperref}
\usepackage{authblk}
\usepackage{soul}

\usepackage{color}

\begin{document}
\title{ModTGCN: Modularity-aware Graph Neural Networks for Text Classification}

\author{Rajarshi Misra\inst{1} \and
Aditya Sharma\inst{1} \and
Vinti Agarwal\inst{1}\thanks{Corresponding author: vinti.agarwal@pilani.bits-pilani.ac.in}
Hari Om Aggrawal\inst{2}}
\authorrunning{Rajarshi Misra, Aditya Sharma, Vinti Agarwal, Hari Om Aggrawal}
%
\institute{ BITS Pilani, India \email{{f20201822p@alumni.bits-pilani.ac.in, \{p20200470,vinti.agarwal\}@pilani.bits-pilani.ac.in}}\and Independent Researcher
\email{hariom85@gmail.com}}
\maketitle              
\begin{abstract}
\sloppy 
{Graph-based text classification models typically rely on local neighborhood aggregation and overlook global community structure, despite semantic document graphs exhibiting strong class-consistent clustering. Ignoring this can blur class boundaries and lead to over-smoothing. We propose ModTGCN, a modularity-aware graph neural network for text classification that jointly optimizes cross-entropy and a modularity-based auxiliary objective to promote class-coherent document communities while preserving discriminative representations. The modularity term is computed on a document–document similarity graph derived from transformer embeddings (pretrained or fine-tuned). To improve scalability, we decouple the original heterogeneous TextGCN graph into separate document–word and word–word components, achieving 2$\times$–10$\times$ faster training. We further study graph construction strategies, label-aware edge reweighting, and supervision choices for modularity optimization. Experiments on five benchmarks show consistent gains, with larger improvements on complex, low homophily  datasets such as Ohsumed and 20NG. }
\end{abstract}

\section{Introduction}
{Text classification remains a fundamental task in natural language processing. 
Recent advances in transformer-based models such as BERT and large language models (LLMs) have achieved strong zero- and few-shot performance \cite{LLMFC,LLMTextClass}. 
However, these models often require costly full fine-tuning, parameter-efficient adaptation (e.g., adapters, LoRA), or prompt calibration to perform well in supervised settings. 
An alternative line of work formulates text classification as \ul{learning over graph structures}, where words and documents are represented as nodes, and edges encode lexical or semantic relationships. 
Graph neural network (GNN) approaches such as TextGCN\cite{textgcn}, TensorGCN\cite{tensorgcn}, BertGCN\cite{BertGCN}, and VGCNBert\cite{VGCNBert}  explicitly model interactions among words and documents, enabling semi-supervised learning through relational propagation.}
{Despite their effectiveness, most graph-based text classifiers rely primarily on local neighborhood aggregation. 
Yet semantic document graphs often exhibit clear mesoscopic structure: \ul{documents sharing the same label tend to form assortative clusters} with dense intra-class connectivity and sparse inter-class links. 
When this global community structure is ignored, two issues commonly arise: (i) hub-driven shortcuts from high-frequency terms or noisy similarities blur class boundaries, and (ii) over-smoothing in deeper GNNs homogenizes representations across weak community separations, reducing discriminability. 
Incorporating global structural information during training may therefore improve robustness and class separation~\cite{connecting}}.

Modularity optimization—originally developed for community detection~\cite{newman2006,Newman16}
—provides a principled mechanism for modeling such global structure. 
Motivated by this observation, we introduce \textbf{ModTGCN}, a modularity-aware GNN for text classification. 
{Our approach augments standard cross-entropy supervision with a modularity-based auxiliary objective computed on a document–document similarity graph derived from transformer embeddings. 
The central hypothesis is that explicitly optimizing modular structure aligns learned representations with class-level graph communities, thereby providing global regularization beyond local message passing. 
Although modularity-aware GNNs have shown promise in other domains~\cite{Murata_2018,tsitsulin}, their use in semi-supervised document classification remains underexplored.}

{A key challenge in semi-supervised graph learning is propagating supervision from limited labeled nodes while respecting global structure. 
We address this by computing modularity on a document–document graph constructed from pretrained or fine-tuned SBERT embeddings~\cite{SBERT}, using gold labels for labeled nodes and TextGCN~\cite{textgcn} predictions for unlabeled nodes.} 
This hybrid supervision scheme yields label-efficient and interpretable improvements without requiring expensive LLM fine-tuning, while remaining encoder-agnostic and compatible with future embedding advances.

{To further improve scalability, we decouple the original single heterogeneous TextGCN graph into separate document–word (TF–IDF) and word–word (PMI) components. 
This preserves the original propagation mechanism while substantially reducing computational overhead on large datasets. 
A detailed complexity analysis is provided in Section~\ref{sec:complexity}.}

To this end, we summarize our contributions as:
\begin{itemize}
    \item \textbf{Modularity-aware GNN with hybrid supervision.} We introduce a joint objective $L = \mathrm{CE} + \lambda(-Q)$ that promotes class-consistent communities on a document–document graph built from language model embeddings, using pseudo-labels for unlabeled nodes.
    \item \textbf{Architectural decoupling of TextGCN.} We reformulate the heterogeneous graph into separate document–word and word–word components, improving scalability without altering the effective decision function.
    \item \textbf{Document–document adjacency strategies.} We compare TF–IDF, cosine, and Gaussian similarity graphs, analyzing accuracy–scalability trade-offs.
    \item \textbf{Empirical validation.} Experiments on five benchmark datasets demonstrate that jointly optimizing classification and modularity consistently improves performance, particularly on structurally complex datasets.
\end{itemize}

\section{Related Work}
Existing graph-based text classification methods can be categorized into \textbf{GNN-only} and \textbf{hybrid GNN–language model (LM)} approaches. \textbf{GNN-only} methods such as TextGCN\cite{textgcn} construct heterogeneous document–word graphs using TF–IDF and PMI edges and apply a two-layer GCN for semi-supervised text classification. Subsequent variants extend this design by incorporating multi-view or tensor-based word graphs\cite{tensorgcn}, heterogeneous attention mechanisms\cite{HeteGCN}, or additional lexical nodes (e.g., character and n-grams) to enrich structural information\cite{wctextgcn}. While effective, these methods rely on large, fine-grained graphs, which incur high computational costs and limit scalability. \textbf{Hybrid GNN–LM } models integrate contextual embeddings from pretrained transformers into graph learning. Approaches such as BERTGCN\cite{BertGCN} and VGCN-BERT\cite{VGCNBert} jointly train or fuse language models with graph encoders to combine local contextual signals and global relational structure. Although they improve performance, co-training LMs with GNNs incurs substantial computational overhead.

\noindent\textbf{Modularity in GNNs.}
Modularity \cite{newman} measures community quality by comparing observed within-community edges against a degree-preserving null model.
A known limitation is the \emph{resolution limit} \cite{resolution-limit}, which causes small but coherent communities to be merged into larger ones. To alleviate this, resolution-adjusted variants introduce a tunable parameter \cite{gamma} (with $\gamma>1$ revealing smaller groups) or adopt \emph{modularity density} \cite{mod-density} and its generalization $Q_g$ \cite{gen-mod-density}, which weights communities by internal link density to better retain cohesive clusters.
Several works integrate modularity into neural objectives:
Modularity regularizers have been added to GAE/VGAE for unsupervised community detection \cite{mod-gnn}, VGAER \cite{Qiu}; Murata \& Afzal \cite{Murata_2018} directly optimize modularity during GNN training, producing embeddings aligned with community structure for clustering tasks. These studies show that modularity-aware learning can uncover latent structure and improve downstream quality.
\emph{\textbf{Connection to our setting:}}
{Prior modularity-aware approaches primarily target unsupervised clustering or community detection. In contrast, we integrate modularity into semi-supervised text classification, aligning graph communities with class labels through hybrid supervision on a document–document similarity graph. This provides a global structural prior that complements local message passing while maintaining computational efficiency via our decoupled graph construction.}

\section{Proposed Method}
\subsection{Problem Formulation}
Given a corpus of documents $\mathcal{D} = \{d_1, d_2, \ldots, d_n\}$ and a corresponding label set $\mathcal{Y} = \{y_1, y_2, \ldots, y_n\}$ spanning $\mathcal{C}$ classes, the primary goal is to perform document classification. During training, document set \(\mathcal{D}\) is partitioned into labeled $U$ and unlabeled $V$ sets, and the objective is to learn a mapping function $f: \mathcal{D} \rightarrow \mathcal{Y}$ that can accurately predict the class $y_i$ for each unlabeled documents {using graph-based relational structure.}

\subsection{ModTGCN: Graph Construction}
\label{sec:graph_const}
{ModTGCN operates over three graphs: (1) a \textit{document--word graph} $\mathcal{G}_d = (\mathcal{V}_d, \mathcal{E}_d)$ constructed via TF--IDF weights, 
(2) \textit{a word--word graph }$\mathcal{G}_w = (\mathcal{V}_w, \mathcal{E}_w)$, built from PMI-based word co-occurrence, and (3) \textit{document-document similarity graph} for modularity optimization, $\mathcal{G}_{doc} = (\mathcal{V}_{doc},\mathcal{E}_{doc})$, where edge weights represent node similarity computed using Gaussian (RBF) kernel $S_{ij} = \exp\!\left(-\tfrac{\|e_i - e_j\|^2}{2\sigma^2}\right)$\cite{lsa}, $e_i$ is the embedding of document $i \in \mathcal{V}_{doc}$ from transformers and $\sigma$ controls the neighborhood sensitivity in kernel space. The first two graphs preserve TextGCN’s propagation structure, while the third introduces global structural supervision.} 

\subsection{Modularity as an Objective Function} 
{Standard GCNs\cite{kipf} aggregate over L-hop local neighborhoods, modularity introduces global, degree-aware coupling between all node pairs via the null-model term. However, semantic document graphs often exhibit mesoscopic community structure aligned with class labels. To explicitly encourage such structure, we introduce modularity as an auxiliary objective.} 
{Modularity, $Q$ measures the deviation of observed intra-community connectivity from a degree-preserving null model.For the label (or community) matrix $P\in\mathbb{R}^{n\times C}$, the modularity is:}

\begin{equation}
\label{eq:modularity}
Q(P) \;=\; \frac{1}{2m}\,\mathrm{Tr}\!\big(P^\top B P\big),
\qquad
B \;=\; A \;-\; \gamma\, \frac{k k^\top}{2m},
\end{equation}
where $A$ is the adjacency, $k$ the degree vector, $m$ the total edge weight, and $\gamma$ the resolution parameter. Modularity matrix $B$ quantifies how much the graph can deviate from the null model.
Building on this, the modularity loss is: 
\begin{equation}
\label{eq:lmod}
\mathcal{L}_{mod}(P) \;=\; -\,Q(P),
\end{equation}
which is minimized when the predicted communities (induced by $P$) have intra-community edge count exceeding the null-model expectation. Importantly, even non-adjacent nodes ($A_{ij}=0$) contribute via the null-model term $\gamma\, \frac{k k^\top}{2m}$, giving modularity its global coupling effect.

\paragraph{Modularity gradient under hybrid supervision.}
{We partition the label matrix $P$ as $(P_U,P_V)$ and the modularity matrix $B$ into blocks $\{B_{UU},B_{UV},B_{VU},B_{VV}\}$ based on labeled, $U$ and unlabeled $V$ nodes.

Modularity $Q$, is computed on $\mathcal{G}_{doc}$, using gold labels for $P_U$ and TextGCN pseudo labels for $P_V$.{%
We also evaluate a variant that uses soft labels for both $P_U$ and $P_V$; see ablation study in Section~\ref{sec:ablation_study}.}}
The gradient with respect to \ $P_V$ is:
\begin{equation}
\label{eq:grad-V}
\nabla_{P_V}\mathcal{L}
\;=\;
-\,\frac{1}{m}\Big(B_{VV} P_V \;+\; B_{VU} P_U\Big).
\end{equation}
 
{The term $B_{VU}P_U$ acts as a degree-corrected supervision field induced by labeled nodes, while $B_{VV}P_V$ couples unlabeled nodes to one another.}
Expanding the supervision field for node $i\in V$ and class $c$:
\begin{equation}
\big[B_{VU}P_U\big]_{i,c}
\;=\;
\sum_{j\in U} A_{ij}\,\mathbf{1}[g_j=c]
\;-\;
\frac{k_i}{2m}\sum_{j\in U} k_j\,\mathbf{1}[g_j=c].
\end{equation}

{Thus, nodes are encouraged toward classes where observed connectivity exceeds null-model expectation, mitigating hub bias and discouraging degenerate single-cluster assignments.}

\begin{figure}[t]
\centering
\begin{subfigure}[t]{0.20\linewidth}
\centering
\resizebox{\linewidth}{!}{%
\begin{tikzpicture}[>=stealth, node distance=1.6cm, every node/.style={transform shape}]
  \tikzstyle{A}=[circle, draw, fill=blue!15, thick, minimum size=7mm]
  \tikzstyle{B}=[circle, draw, fill=red!15, thick, minimum size=7mm]
  \tikzstyle{U}=[circle, draw, fill=gray!10, thick, minimum size=7mm]

  \node[A] (1) {1};
  \node[U, above=of 1] (2) {2};
  \node[B, right=of 2] (3) {3};
  \node[A, right=of 1, yshift=-2cm] (4) {4};
  \node[U, right=of 4, yshift=2cm] (5) {5};
  \node[U, right=of 3] (6) {6};

  \draw (1) -- (2);
  \draw (4) -- (2);
  \draw (3) -- (6);
  \draw (4) -- (5);
  \draw (4) -- (6);
  \draw (5) -- (6);
\end{tikzpicture}%
}
\caption{S1: baseline.}
\end{subfigure}\hfill
\begin{subfigure}[t]{0.20\linewidth}
\centering
\resizebox{\linewidth}{!}{%
\begin{tikzpicture}[>=stealth, node distance=1.6cm, every node/.style={transform shape}]
  \tikzstyle{A}=[circle, draw, fill=blue!15, thick, minimum size=7mm]
  \tikzstyle{B}=[circle, draw, fill=red!15, thick, minimum size=7mm]
  \tikzstyle{U}=[circle, draw, fill=gray!10, thick, minimum size=7mm]

  \node[A] (1) {1};
  \node[U, above=of 1] (2) {2};
  \node[B, right=of 2] (3) {3};
  \node[A, right=of 1, yshift=-2cm] (4) {4};
  \node[U, right=of 4, yshift=2cm] (5) {5};
  \node[U, right=of 3] (6) {6};

  \draw (1) -- (2);
  \draw (4) -- (2);
  \draw (3) -- (6);
  \draw (4) -- (5);
  \draw (4) -- (6);
  \draw (5) -- (6);

  \draw[very thick] (4) -- (1);
  \draw[very thick] (4) -- (3);
  \draw[very thick] (1) -- (5);
\end{tikzpicture}%
}
\caption{S2: global change.}
\end{subfigure}\hfill
\begin{subfigure}[t]{0.20\linewidth}
\centering
\resizebox{\linewidth}{!}{%
\begin{tikzpicture}[>=stealth, node distance=1.6cm, every node/.style={transform shape}]
  \tikzstyle{A}=[circle, draw, fill=blue!15, thick, minimum size=7mm]
  \tikzstyle{B}=[circle, draw, fill=red!15, thick, minimum size=7mm]
  \tikzstyle{U}=[circle, draw, fill=gray!10, thick, minimum size=7mm]

  \node[A] (1) {1};
  \node[U, above=of 1] (2) {2};
  \node[B, right=of 2] (3) {3};
  \node[A, right=of 1, yshift=-2cm] (4) {4};
  \node[U, right=of 4, yshift=2cm] (5) {5};
  \node[U, right=of 3] (6) {6};

  \draw (1) -- (2);
  \draw (4) -- (2);
  \draw (3) -- (6);
  \draw (4) -- (5);
  \draw (4) -- (6);
  \draw (5) -- (6);
  \draw (4) -- (1);
  \draw (4) -- (3);
  \draw (1) -- (5);

  \draw[very thick, dashed] (2) -- (5);
\end{tikzpicture}%
}
\caption{S2+: add (2--5).}
\end{subfigure}

\caption{Toy graphs with labeled nodes $U=\{1,3,4\}$ (blue, red, blue) and unlabeled nodes (grey) $V=\{2,5,6\}$. Node~2 always connects to nodes~1 and~4 (both A). Blue/red = class A/B. Thick edges are added in S2; dashed in S2+.}

\label{fig:toy}
\end{figure}
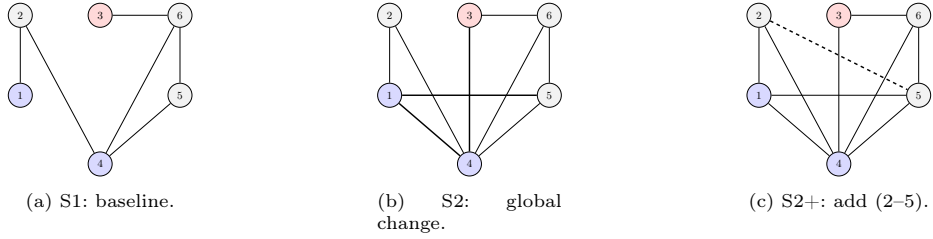

\paragraph{Global coupling in modularity \(Q\) (toy).}
Figure~\ref{fig:toy} shows three cases highlighting the global nature of modularity. In \textit{S1 (baseline)}, node~2 links to moderate-degree blue nodes. {Because these observed edges exceed the degree-corrected null-model expectation, the supervision field favors the blue class and penalizes red.} 
{In \textit{S2 (global change)}, adding $(1,5)$ and $(3,4)$ increases blue node degrees and $m$ without altering node~2’s neighborhood, raising the null-model baseline and weakening the blue field, despite its immediate connections remaining unchanged. This demonstrates the dependence of modularity on the \emph{global} degree distribution, rather than local adjacency.} 
In \textit{S2+}, adding $(2,5)$ changes $k_2$ and $m$ and activates $B_{VV}P_V$ in \eqref{eq:grad-V}, so node~2 is not only influenced by labeled neighbors but also by unlabeled neighbors’ soft labels (e.g., a red-leaning node~5 reduces its preference for blue), illustrating unlabeled--unlabeled coupling.

\paragraph{Observation.}
This mechanism provides (i) \emph{global supervision propagation} via the degree-corrected field $B_{VU}P_U$, allowing few labels to influence distant nodes, and (ii) \emph{unlabeled--unlabeled coherence} via $B_{VV}P_V$, aligning predictions among unlabeled neighbors. {Overall, modularity propagates supervision through a degree-aware global field: labeled nodes influence distant regions via the null model, and unlabeled nodes align through mutual coupling. This mechanism complements local GCN aggregation by enforcing mesoscopic community consistency while mitigating hub-driven shortcuts and degenerate single-cluster solutions.}

\subsubsection{Joint Optimization Objective}
For decoupled-TextGCN, we use the standard cross-entropy loss on labeled nodes:
\begin{equation}
\mathcal{L}_{CE} \;=\; - \mathrm{Tr}\!\left(Y_U^\top \log \hat{Y}_U \right),
\end{equation}
where $\mathrm{Tr}(\cdot)$ is the trace, $Y_U$ is the one-hot label matrix, and $\hat{Y}_U$ denotes the predicted label distributions. Predictions are computed as
\begin{equation}
\label{eq:ypred}
\hat{Y} = \mathrm{softmax}\!\left( \mathcal{A}_{dw} 
\left( \mathcal{A}_{dw}^\top W_1 + \mathcal{A}_{ww} W_2 \right) W_3 \right),
\end{equation}
with $\mathcal{A}_{dw}$ and $\mathcal{A}_{ww}$ the document--word and word--word adjacencies, and $W_1,W_2$, $W_3$ trainable weights. This preserves the original TextGCN propagation while improving scalability (see Section~\ref{sec:complexity}).
The final objective combines classification and modularity:
\begin{equation}
\mathcal{L}_{total} = \lambda \mathcal{L}_{CE} - (1-\lambda)\mathcal{L}_{mod},
\end{equation}
where $\lambda\in[0,1]$ controls the trade-off between predictive accuracy and community-structure preservation.

\section{Experimental setup}
\subsection{Baselines}
\label{sec:baselines}
We evaluate our method on five benchmark datasets: MR, R8, R52, 20 Newsgroups (20NG), and Ohsumed as used in TextGCN \cite{textgcn}. {A detailed statistics of datasets are available on our website \cite{website}.} We compare our framework against three broad categories of models. \textbf{GNN-based models:} {TextGCN} \cite{textgcn}, {TensorGCN} \cite{tensorgcn}, {WCTextGCN} \cite{wctextgcn}; \textbf{Classification on BERT embeddings:} Logistic Regression(LR) and Linear Probing (a single-layer MLP); \textbf{LLM in zero and few-shot settings:} \texttt{OpenAI}-GPT-3, ChatGPT; \textbf{Ours}: ModTGCN with $\mathcal{G}_{doc}$ using pre-trained/fine-tuned embeddings.

\subsection{Implementation and Evaluation Metrics}
{
We follow TextGCN\cite{textgcn} configuration, including preprocessing, sliding window size ($20$), dataset splits, and hidden layer size ($200$). The model is trained with Adam optimizer for up to 300 epochs with early stopping (patience 30).} 
For modularity optimization, hyperparameters including dropout, $\sigma$, learning rate $\eta$, resolution $\gamma$, loss weight $\lambda$, and edge-weight parameters $(\alpha,\beta,k)$ with tuned with \texttt{Optuna}.
{Performance is reported using micro-F1, averaged over five random seeds for stability. The code is available on GitHub 
\emph{\url{https://github.com/Rajarshi-Misra/ModTGCN}}}. 
\section {Results}
\subsection{Baseline Performance}
Table~\ref{tab:cmp_models} shows that modularity-based optimization substantially improves performance over TextGCN and related GNN baselines. Using pre-trained embeddings, \textbf{ModTGCN(P)} achieves clear gains over TextGCN across datasets, with particularly strong improvements on MR($+4.7$), Ohsumed($+3.6$) and 20NG($+4.3$ ).
\begin{table}[t]
\centering
\caption{Micro-F1 (mean\% $\pm$ std) of competing methods on MR, R8, R52, Ohsumed and 20NG. Abbreviations: (P)/(F) — pre-trained/fine-tuned SBERT-\texttt{all-mpnet-base} embeddings used to build the doc–doc graph for the modularity term; LR — Logistic Regression.
Best results per dataset are shown in \textbf{bold}. {The homophily values corresponding to each dataset are reported in parentheses.} }
\label{tab:cmp_models}
\scalebox{0.9}{%
\begin{tabular}{@{}p{3.8cm}ccccc@{}}
\toprule
\textbf{Model} & \textbf{MR(0.70)} & \textbf{R8(0.50)} & \textbf{R52(0.38)} & \textbf{Ohsumed(0.16)} & \textbf{20NG(0.19)} \\
\midrule
TextGCN & 76.74$\pm$0.20 & 97.07$\pm$0.13 & 93.56$\pm$0.18 & 68.36$\pm$0.56 & 86.34$\pm$0.09 \\
TensorGCN & 77.91$\pm$0.07 & {98.04$\pm$0.08} & {95.05$\pm$0.11} & 70.11$\pm$0.24 & 87.74$\pm$0.05 \\
WCTextGCN \cite{wctextgcn} & 77.85$\pm$0.34 & 97.49$\pm$0.20 & 93.88$\pm$0.34 & 68.52$\pm$0.20 & 87.21$\pm$0.54 \\
U-TextGCN \cite{han2022understandinggraphconvolutionalnetworks} & 76.41 & 96.92 & 93.45 & 68.24 & 86.07 \\
\midrule
\begin{tabular}[c]{@{}l@{}}BertGCN \cite{BertGCN} (Co-trained \\  BERT with TextGCN)\end{tabular} & 86.0 & 98.1 & 96.6 & 72.8 & 89.5 \\
\midrule
\multicolumn{6}{l}{\textbf{Pre-trained embeddings(P)}} \\
LR(P)& 83.73 & 98.12 & 95.63 & 71.30 & 79.06 \\
Linear Probing(P)& 83.49$\pm$0.09 & 98.25$\pm$0.07 & 95.26$\pm$0.05 & 69.87$\pm$0.09 & 79.06$\pm$0.14 \\
\textbf{\begin{tabular}[c]{@{}l@{}}ModTGCN(P)\end{tabular}} & {81.45$\pm$0.05} & \text{97.55$\pm$0.06} & \text{94.54$\pm$0.11} & {71.97$\pm$0.15} & {90.6 ± 0.1}\\

\midrule
\multicolumn{6}{l}{\textbf{Fine-tuned embeddings(F)}} \\
LR(F)& 87.87 & 98.04 & \textbf{96.73} & 74.65 & 87.17 \\
{Linear Probing(F)}& 87.37$\pm$0.20 & 98.10$\pm$0.12 & 96.71$\pm$0.11 & 74.76$\pm$0.17 & 86.90$\pm$0.10 \\
\textbf{\begin{tabular}[c]{@{}l@{}}ModTGCN(F)\end{tabular}} & \textbf{88.07$\pm$0.06} & \textbf{98.70$\pm$0.09} & \text{96.16$\pm$0.16} & \textbf{77.52$\pm$0.25} & \textbf{91.14 $\pm$ 0.12}\\
\hline
\text{\begin{tabular}[c]{@{}l@{}}{ModTGCN(F)}\\ {warmup}\end{tabular}} & \text{88.04} & \text{98.40} & \text{96.14} & \text{76.34} & \text{89.91}\\
\bottomrule
\end{tabular}}
\end{table}
{To examine the complementarity of modularity with domain-adapted representations, we fine-tuned \texttt{SBERT} on each dataset before constructing the document graph. Although LR(F) and Linear Probing(F) benefit from fine-tuning, \textbf{ModTGCN(F)} achieves the highest micro-F1 on most datasets, with substantial improvements over its pre-trained counterpart on MR ($+6.6$) and Ohsumed ($+5.5$).} Moreover, despite its simpler architecture, ModTGCN matches TensorGCN's performance on R8 ($97.5$ vs. $98.0$) and comes within $0.5$ points on R52.
{While overall improvements are consistent, the magnitude of gains varies across datasets. We therefore analyze the structural conditions under which modularity is most beneficial. Compared to embedding-only baselines (LR(P) and Linear Probing(P)), ModTGCN achieves slightly lower performance on simpler, high-homophily datasets (MR $-2.05$, R8 $-0.70$, R52 $-0.72$; homophily $0.70$, $0.50$, $0.38$), where pretrained representations are already near linearly separable, leaving limited room for structural refinement. In contrast, modularity yields substantial gains on structurally complex, low-homophily datasets (Ohsumed $+2.10$, 20NG $+11.54$; homophily $0.16$, $0.19$). This trend indicates that modularity refinement is most beneficial in low-homophily regimes with overlapping semantic boundaries, and less impactful on trivially separable datasets.}
{Overall, these results demonstrate that modularity-based optimization remains effective and complementary under both pre-trained and fine-tuned embedding settings.}

\subsection{Comparison with LLMs}
Table \ref{tab:llm} compares ModTGCN against zero-shot and few-shot LLM baselines. Reported GPT-3 and ChatGPT scores are taken directly from their respective papers, and missing entries indicate datasets not evaluated in those works. Both \textbf{ModTGCN(P)} and \textbf{ModTGCN(F)} outperform  LLM variants. {The advantage is most evident on Ohsumed and 20NG, suggesting that explicit graph-structured supervision can outperform prompting-based inference when labeled graph structure is available.}

\begin{table*}[t]
\centering
\begin{minipage}[t]{0.54\textwidth}
    \centering
    \caption{Comparison of LLM baselines with ModTGCN. GPT-3 uses \texttt{InstructGPT-3 (text-davinci-003)}.}
    \label{tab:llm}
    \scalebox{0.78}{
    \begin{tabular}{lccccc}
        \toprule
        Model & MR & R8 & R52 & Ohsumed & 20NG \\
        \midrule
        \multicolumn{6}{l}{\textbf{LLM Zero-shot}} \\
        GPT-3  & 88.69 & 90.19 & 89.06 & -- & -- \\
        ChatGPT & -- & 60.10 & 75.23 & 39.93 & 58.70 \\
        \midrule
        \multicolumn{6}{l}{\textbf{LLM Few-shot (k)}} \\
        GPT-3 (k=16) & \textbf{89.15} & 91.58 & 91.56 & -- & -- \\
        ChatGPT (k=2) & -- & 72.54 & 81.68 & 47.05 & 58.44 \\
        ChatGPT (k=5) & -- & 82.43 & 90.13 & 45.39 & -- \\
        \midrule
        ModTGCN(P) & 81.45 & 97.55 & 94.54 & 71.97 & 90.60 \\
        ModTGCN(F) & 88.07 & \textbf{98.70} & \textbf{96.14} & \textbf{77.52} & \textbf{91.14} \\
        \bottomrule
    \end{tabular}}
\end{minipage}
\hfill
\begin{minipage}[t]{0.42\textwidth}
    \centering
    \caption{Training time (seconds) and epochs for decoupled vs. original TextGCN. Decoupling reduces per-epoch cost, yielding $2\times– 10\times$ speedup.}
    \label{tab:complexity}
    \scalebox{0.85}{
    \begin{tabular}{lcccc}
        \toprule
        Dataset & \multicolumn{2}{c}{Decoupled} & \multicolumn{2}{c}{Original} \\
        \cmidrule(lr){2-3} \cmidrule(lr){4-5}
         & Time (s) & Epochs & Time (s) & Epochs \\
        \midrule
        MR      & \textbf{10.50} & 111 & 11.50 & 16 \\
        R8      & \textbf{36.90} & 191 & 87.60 & 97 \\
        R52     & \textbf{64.55} & 227 & 257.60 & 163 \\
        Ohsumed & \textbf{80.62} & 294 & 241.10 & 81 \\
        20NG    & \textbf{149.60} & 243 & 1450.40 & 111 \\
        \bottomrule
    \end{tabular}}
\end{minipage}
\end{table*}

\subsection{Complexity and scalability analysis }
\label{sec:complexity}
{The original TextGCN applies two-hop propagation over a single heterogeneous adjacency \(A = [0, \mathcal{A}_{dw}; \mathcal{A}_{wd}, \mathcal{A}_{ww}]\), evaluating both document and word branches at each hop. This incurs sparse cost of \(O((4E_{dw}+2E_{ww})H)\) and unnecessary word-node logit computation.}
In contrast, our decoupled approach follows only the \emph{document-logit} flow (Eq.~\eqref{eq:ypred}) reducing sparse operations to \(O((2E_{dw}+E_{ww})H)\). {This effectively reduces \(\approx 50\%\) edge traversals per layer.}
{The decoupled formulation is lossless for document classification, as it is directly derived from the original TextGCN equations. Since loss is computed only on document logits, removing the unused word branch preserves the decision function while eliminating redundant computation. A detailed derivation is added in section \ref{appendix:computation} of the Appendix.}

Empirically (Table \ref{tab:complexity}), the decoupled variant achieves $2\times$ to $10\times$ faster training than the original TextGCN despite using more epochs, indicating much lower per-epoch cost. The improvement is most pronounced on 20NG, where training time drops from $1,450$ seconds to $150$ seconds, confirming improved scalability. 

\section{Ablation studies}
\label{sec:ablation_study}
\subsection{Effect of graph-style, edge reweighting, and labeling strategy}
{Table \ref{tab:graph_results} evaluates three factors affecting modularity training: (i) $\mathcal{G}_{doc}$ construction method; (ii) label-aware edge reweighting; and (iii) supervision source for modularity.}

\textbf{Graph-style (Gr):} We evaluate three $G_{doc}$ construction strategies: 
(i) \textit{TF–IDF inner product} (tf): $S_{ij} = (A_{dw}A_{dw}^\top)_{ij}$, 
where $A_{dw}$ is the TF–IDF matrix. This approach does not use language models and relies purely on sparse matrix multiplication over lexical features. 
(ii) \textit{Cosine similarity} (c): $S_{ij} = \cos(e_i, e_j)$, where $e_i$ is the embedding of the $i^{th}$ sentence. (iii) \textit{Gaussian similarity} (g): $S_{ij} = \exp\!\left(-\frac{\|e_i - e_j\|^2}{2\sigma^2}\right)$; see Section~\ref{sec:graph_const}. 
The latter two approaches construct the document graph using transformer-based embeddings.
{\noindent\textit{Observations:} Across all datasets, Gaussian (RBF) graphs consistently outperform cosine- and TF--IDF-based similarities, indicating localized kernel affinities better capture semantic neighborhoods. 
Moreover, among the three encoders in Table \ref{tab:ablation_lang},  
\texttt{SBERT-all-mpnet-\allowbreak base} achieves the best overall accuracy and is the most consistent across datasets. } 

\textbf{Edge reweighting (W):} To inject label supervision into the modularity, we amplify same-class edges and attenuate cross-class edges by factors $\alpha>1$ and $\beta<1$ over each training document’s top-$k$ neighbors. Edges connecting validation/test nodes keep their original weights to prevent label leakage. This retains semantic similarity while promoting label-consistent communities.
\noindent{\textit{Observation:} Enabling label-aware edge reweighting improves performance by strengthening intra-class similarities, suggesting that modest supervision injected at the similarity level enhances modular community formation.}

\textbf{Labeling strategy (L):} We evaluate two supervision choices for the modularity loss: (i) gold labels on training nodes with predicted labels on validation/test nodes (gt), and (ii) predicted labels for \emph{all} nodes (p). 
\noindent\textit{Observations:} Using predictions (p) in place of ground-truth labels ($\mathbf{gt}$) often improves performance, particularly under class imbalance. On Ohsumed, for example, Gaussian + Edge-reweighting + Pred achieves $72.0/65.0$ vs.\ $71.5/63.3$ with gold training labels. This indicates that prediction-based supervision can regularize modularity optimization and reduce overfitting to a limited set of labeled nodes. 

Overall, ModTGCN under \{Gaussian, reweighting enabled, predicted labels\}(\{g,T,p\}) configuration gives the strongest results. We use the same configuration for all other further experiments. {Table~4 also shows that using predicted labels for all nodes performs slightly better than gold-only supervision, indicating robustness to pseudo-label errors. Moreover, modularity increases smoothly over epochs (Figure \ref{fig:cevsmod}), suggesting stable community refinement rather than error amplification. A warm-up strategy (ModTGCN(F) warmup), delaying modularity until few epochs in Table~\ref{tab:cmp_models}, yields only marginal differences($-0.03$ on MR, $-0.30$ on R8, $-0.02$ on R52, $-1.18$ on Ohsumed, $-1.23$ on 20NG), confirming that early prediction noise does not significantly affect convergence.}

\begin{table}[t]
\centering
\caption{Micro/Macro F1 performance (mean ± std) under different configurations. Graph style: Gr $\in$ ({c,tf,g});(c=Cosine, tf=TF--IDF, g=Gaussian), Edge reweighing: W $\in$ ({T, F}); T - if weighted adjustment is applied, else F, and Labeling strategy: L $\in$ ({gt,p}); gt \& p denote ground-truth and predicted labels for train-set respectively.}
\label{tab:ablation_comb}
\resizebox{\linewidth}{!}{%
\begin{tabular}{lllccccl}
\toprule
Gr & W & L & MR & R8 & R52 & Ohs  &20NG\\
\midrule
c  & T & gt & 81.2 ± 0.1 / 81.2 ± 0.1 & 97.4 ± 0.1 / 92.2 ± 0.4 & 94.1 ± 0.2 / 67.0 ± 1.5 & 71.5 ± 0.3 / 63.3 ± 0.5  &\textbf{91.1 ± 0.1 / 90.5± 0.1}\\
c  & T & p & 81.4 ± 0.2 / 81.4 ± 0.2 & 97.4 ± 0.2 / 92.2 ± 0.3 & 94.3 ± 0.2 / 69.1 ± 2.4 & 71.7 ± 0.2 / 65.0 ± 0.4  &90.8 ± 0.2 / 90.1 ± 0.2\\
c  & F & gt & 81.6 ± 0.1 / 81.6 ± 0.1 & 97.1 ± 0.1 / 92.0 ± 0.5 & 93.7 ± 0.2 / 69.0 ± 2.0 & 68.9 ± 0.3 / 61.9 ± 0.7  &88.2 ± 0.1 / 87.3 ± 0.2\\
c  & F & p & 80.8 ± 0.1 / 80.8 ± 0.1 & 96.8 ± 0.1 / 90.5 ± 2.3 & 93.9 ± 0.1 / 67.6 ± 1.2 & 68.2 ± 0.9 / 60.8 ± 2.1  &88.1 ± 0.2 / 87.3 ± 0.2\\
\midrule
tf & T & gt & 76.4 ± 0.3 / 76.4 ± 0.3 & 95.8 ± 0.3 / 86.6 ± 1.9 & 92.9 ± 0.1 / 66.6 ± 2.4 & 68.3 ± 0.3 / 62.6 ± 0.2  &86.9± 0.1/ 86.2 ± 0.1\\
tf & T & p & 76.6 ± 0.2 / 76.6 ± 0.2 & 96.3 ± 0.2 / 88.5 ± 2.2 & 92.8 ± 0.1 / 66.9 ± 2.2 & 68.2 ± 0.3 / 62.5 ± 0.4  &86.7 ± 0.2 / 86.2 ± 0.2\\
tf & F & gt & 77.0 ± 0.1 / 77.0 ± 0.1 & 97.0 ± 0.2 / 91.7 ± 1.7 & 93.5 ± 0.3 / 66.8 ± 3.9 & 68.1 ± 0.1 / 62.4 ± 0.3  &86.5 ± 0.2 / 86.0 ± 0.3\\
tf & F & p & 76.8 ± 0.1 / 76.8 ± 0.1 & 97.0 ± 0.2 / 90.8 ± 1.5 & 93.9 ± 0.1 / 69.4 ± 1.8 & 67.9 ± 0.6 / 61.3 ± 1.0  &86.3 ± 0.3 / 86.0 ± 0.2\\
\midrule
g  & T & gt & 81.6 ± 0.0 / 81.6 ± 0.0 & \textbf{97.6 ± 0.3 / 92.7 ± 1.2} & 94.5 ± 0.2 / 67.6 ± 1.9 & 71.5 ± 0.3 / 63.3 ± 0.7  &90.7 ± 1.0 / 90.0 ± 0.9\\
g  & T & p & 81.4 ± 0.1 / 81.4 ± 0.1 & 97.5 ± 0.1 / 92.3 ± 0.4 & \textbf{94.5 ± 0.1 / 69.9 ± 3.0} & \textbf{72.0 ± 0.1 / 65.0 ± 0.4}  &90.6 ± 0.1 / 89.9 ± 0.1\\
g  & F & gt & \textbf{82.4 ± 0.1 / 82.4 ± 0.1} & 97.0 ± 0.2 / 90.2 ± 1.8 & 93.9 ± 0.2 / 67.7 ± 2.0 & 70.7 ± 0.2 / 63.9 ± 0.5  &90.0 ± 0.1 / 89.1 ± 0.1\\
g  & F & p & 81.1 ± 0.1 / 81.1 ± 0.1 & 96.7 ± 0.3 / 90.0 ± 1.6 & 93.8 ± 0.2 / 67.2 ± 1.6 & 69.2 ± 0.3 / 62.9 ± 0.3  &90.0 ± 0.1 / 89.2 ± 0.1\\
\bottomrule
\end{tabular}
}
\label{tab:graph_results}
\end{table}

\begin{table}[t]
    \centering
    \caption{Performance of ModTGCN under \{g,T,p\} configuration across three pre-trained transformer embeddings.}
    \label{tab:ablation_lang}
    \scalebox{0.95}{
    \begin{tabular}{llllll}
    \toprule
    Dataset & MR & R8 & R52 & Ohs  &20NG\\
    \midrule
    \texttt{SBERT-all-mpnet-base} & \textbf{81.45$\pm$0.05} & \textbf{97.55$\pm$0.06} & \textbf{94.54$\pm$0.11} & \textbf{71.97$\pm$0.15} & \textbf{90.60$\pm$0.10} \\
    \texttt{BERT-base-uncased} & 76.89$\pm$0.41 & 96.76$\pm$0.46 & 93.72$\pm$0.18 & 67.72$\pm$0.12 & 86.85$\pm$0.12 \\
    \texttt{RoBERTa-large} & 77.12$\pm$0.52 & 96.99$\pm$0.21& 93.90$\pm$0.15 & 68.10$\pm$0.21 & 86.93$\pm$0.09\\
    \bottomrule
    \end{tabular}}
\end{table}
\subsection{Hyperparameter sensitivity analysis}
Table~\ref{tab:ablation} analyzes sensitivity to key parameters modularity resolution $\gamma$, loss weight$\lambda$, reweighting factors $\alpha, \beta$, and neighborhood size $k$ on ModTGCN. The loss weight $\lambda$ performs best at intermediate range \{$0.25-0.5$\}, indicating that classification and modularity objectives should be balanced. Similarly, moderate value \{$\sim 3.0$\} for resolution parameter $\gamma$ yield optimal values, indicating that neither overly coarse nor overly fine communities are optimal. Edge reweighting factors $\alpha,\beta$   and neighborhood size $k$ also show clear optima: smaller $\alpha,\beta$ \{$\alpha \sim 1.2, \beta \sim 0.4$\} with a moderate $k$\{$\sim 10$\} produce the strongest results. These results indicate that the benefits of modularity optimization are consistent and robust across reasonable parameter choices.
\begin{table*}[t]
\centering
\caption{Ablation studies for key hyperparameters $\gamma, \lambda, \alpha, \beta$, and $k$. Micro-F1 is reported in (\%).}
\label{tab:ablation}
\scriptsize

\begin{minipage}{0.32\textwidth}
\centering
\scalebox{0.85}{
\begin{tabular}{c|ccccc}
\toprule
$\gamma$  & MR & R8 & Ohsumed & R52 & 20NG\\
\midrule
0.1 & 75.7 & 97.1 & 72.6 & 94.4 & 87.8 \\
0.5 & 80.0 & 97.4 & 69.5 & 94.4 & 87.8 \\
1.0 & 80.6 & \textbf{97.5} & 70.7 & \textbf{94.5} & 88.6 \\
3.0 & \textbf{81.6} & 97.3 & \textbf{71.9} & 94.4 & \textbf{89.9} \\
5.0 & 81.4 & 97.3 & \textbf{71.9} & 94.0 & 90.7 \\
\bottomrule
\end{tabular}
}
\end{minipage}\hfill
\begin{minipage}{0.32\textwidth}
\centering
\scalebox{0.85}{
\begin{tabular}{c|ccccc}
\toprule
$\lambda$  & MR & R8 & Ohsumed & R52 & 20NG\\
\midrule
0.0  & 68.4 & 5.5 & 16.5 & 0.7 & 4.6 \\
0.25 & \textbf{81.3} & 93.7 & 67.6 & \textbf{94.5} & \textbf{90.6} \\
0.5  & \textbf{81.3} & 96.8 & 71.3 & 94.1 & 90.4 \\
0.75 & 79.4 & \textbf{97.4 }& \textbf{72.8} & 93.8 & 89.9 \\
1.0  & 76.5 & 96.8 & 66.5 & 93.4 & 86.0 \\
\bottomrule
\end{tabular}
}
\end{minipage}\hfill
\begin{minipage}{0.32\textwidth}
\centering
\scalebox{0.85}{
\begin{tabular}{c|ccccc}
\toprule
$k$ & MR & R8 & Ohsumed & R52 & 20NG\\
\midrule
4  & 79.52 & 97.26 & \textbf{72.57} & \textbf{94.63} & 90.32 \\
6  & 79.38 & \textbf{97.30} & 72.40 & 94.47 & 90.28 \\
8  & \textbf{81.04} & \textbf{97.30} & 72.52 & 94.55 & \textbf{90.45} \\
10 & \textbf{81.04} & 97.44 & 72.42 & 94.61 & 90.04 \\
12 & 80.53 & 97.44 & 72.37 & 94.59 & 89.95 \\
\bottomrule
\end{tabular}
}
\end{minipage}

\begin{minipage}{0.5\textwidth}
\centering
\scalebox{0.85}{
\begin{tabular}{c|ccccc}
\toprule
$\beta$ & MR & R8 & Ohsumed & R52 & 20NG\\
\midrule
0.3 & 80.5 & \textbf{97.21} & \textbf{72.59} & 94.7 & \textbf{90.72} \\
0.4 & \textbf{81.29} & 97.21 & 72.57 & 94.7 & 90.72 \\
0.5 & 81.23 & 97.21 & 72.57 & 94.7 & 90.63 \\
0.6 & 81.20 & 97.21 & 72.57 & 94.7 & 89.87 \\
0.7 & 81.26 & 97.17 & 72.57 & 94.7 & 89.87 \\
\bottomrule
\end{tabular}
}
\end{minipage}\hfill
\begin{minipage}{0.5\textwidth}
\centering
\scalebox{0.85}{
\begin{tabular}{c|ccccc}
\toprule
$\alpha$ & MR & R8 & Ohsumed & R52 & 20NG\\
\midrule
1.0 & 80.84 & 97.30 & 72.57 & 94.31 & \textbf{90.59} \\
1.2 & \textbf{81.06} & \textbf{97.39} & 72.54 & 94.51 & 90.17 \\
1.4 & \textbf{81.06} & \textbf{97.39} & 72.57 & 94.70 & 90.04 \\
1.6 & 81.26 & 97.21 & \textbf{72.59} & \textbf{94.74} & 89.86 \\
1.8 & 80.84 & 97.17 & 72.57 & 94.70 & 89.88 \\
\bottomrule
\end{tabular}
}
\end{minipage}

\end{table*}
\subsection{Optimization dynamics of CE and Modularity}
\begin{figure}[t]
    \centering
    \subfloat[]{%
        \includegraphics[width=0.48\linewidth]{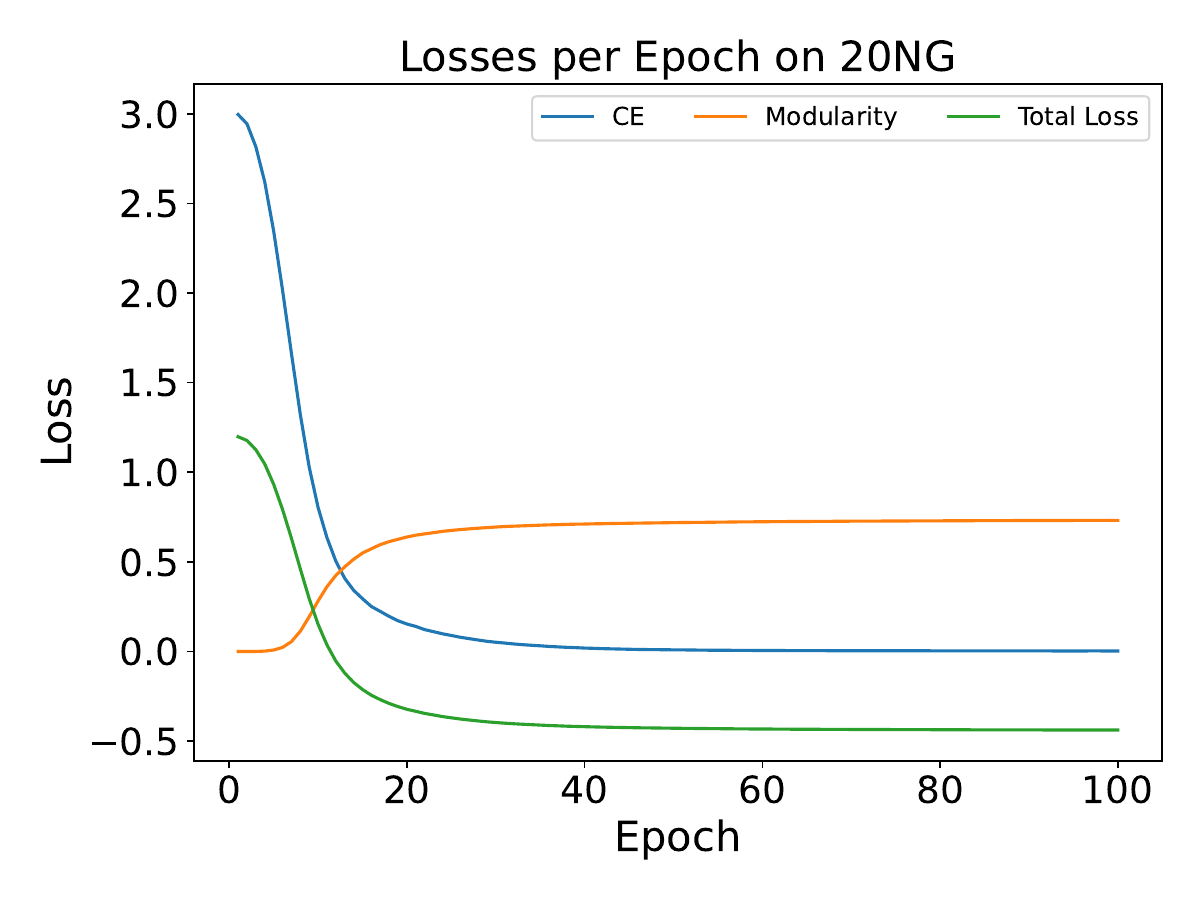}
        \label{fig:cevsmod}   
    }
    \hfill
    \subfloat[]{%
        \includegraphics[width=0.48\linewidth]{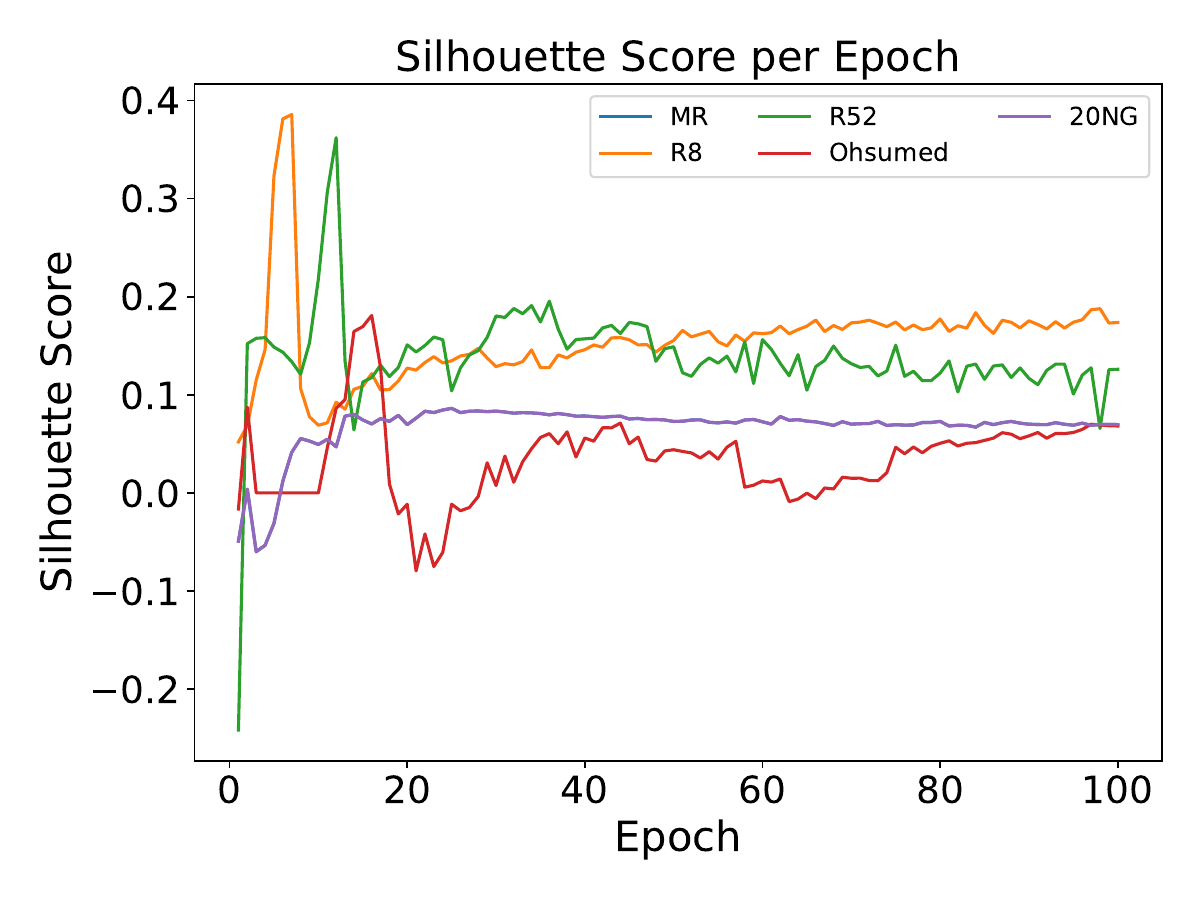}
        \label{fig:silvsepoch}
    }
    \caption{(a) Cross-entropy, modularity, and total loss across epochs on 20NG. (b) Silhouette score across epochs, showing improved cluster separation during training.}
\end{figure}

{Figure~\ref{fig:cevsmod} illustrates that cross-entropy decreases rapidly in early epochs, establishing initial decision boundaries. As predictions stabilize, modularity increases steadily, reinforcing community-level alignment. The smooth convergence of total loss suggests complementary optimization rather than objective conflict. Thus, cross-entropy shapes local classification boundaries, while modularity refines global embedding geometry.
This behavior is further reflected in silhouette scores (Figure \ref{fig:silvsepoch}), which rise and stabilize over epochs, indicating improved intra-class cohesion and inter-class separation.}
\emph{Additional results, plots, dataset details, and detailed derivation of decoupling approach are available on the  website\cite{website}} 
\section{Conclusion}
We introduced a modularity-aware GNN framework that complements the local neighborhood aggregation with a global, community-aware objective for text classification. By integrating modularity into training, the model encourages class-coherent document communities and mitigates degree-related biases that can lead to over-smoothing or trivial partitions.
Experiments across five benchmark datasets demonstrate consistent improvements over baselines. 
The best results are obtained using Gaussian similarity with prediction-based supervision, underscoring the value of global similarity modeling and hybrid label supervision. 
Overall, modularity optimization provides a lightweight yet effective complement to GCN-based text classification. Beyond classification, this framework suggests broader opportunities for incorporating community-aware objectives into graph learning tasks where mesoscopic structure plays a central role, including topic modeling, document clustering, and knowledge graph analysis.

\bibliographystyle{splncs04}
\bibliography{reference}

\appendix
\section{Appendix}
\subsection{Decoupled Formulation of TextGCN}
\label{appendix:computation}
The original TextGCN operates on a heterogeneous graph with adjacency:
\begin{equation}
A =
\begin{bmatrix}
0 & A_{dw} \\
A_{wd} & A_{ww}
\end{bmatrix},
\end{equation}
where $A_{dw} \in \mathbb{R}^{n_d \times n_w}$ denotes document--word edges,
$A_{wd} = A_{dw}^\top$, and $A_{ww} \in \mathbb{R}^{n_w \times n_w}$ encodes
word--word co-occurrence.

Let the node representations at layer $l$ be:
\begin{equation}
H^{(l)} =
\begin{bmatrix}
H_d^{(l)} \\
H_w^{(l)}
\end{bmatrix},
\end{equation}
where $H_d^{(l)}$ and $H_w^{(l)}$ denote document and word embeddings, respectively.

A single GCN layer performs:
\begin{equation}
H^{(l+1)} = A H^{(l)} W.
\end{equation}

Expanding block-wise:
\begin{align}
H_d^{(l+1)} &= A_{dw} H_w^{(l)} W_d, \\
H_w^{(l+1)} &= A_{wd} H_d^{(l)} W_{wd} + A_{ww} H_w^{(l)} W_{ww}.
\end{align}

\paragraph{Eliminating word-node states.}
Since supervision is applied only on document nodes, the word embeddings
$H_w^{(l)}$ act as intermediate variables and can be eliminated via substitution.
Ignoring higher-order recursion, we approximate:
\begin{equation}
H_w^{(l)} \approx A_{wd} H_d^{(l)} W_1 + A_{ww} W_2,
\end{equation}
where the first term captures document-induced signals and the second term
encodes intrinsic word co-occurrence structure.

Substituting into the document update:
\begin{align}
H_d^{(l+1)} 
&= A_{dw} \left( A_{wd} H_d^{(l)} W_1 + A_{ww} W_2 \right) W_3.
\end{align}

Using $A_{wd} = A_{dw}^\top$, we obtain:
\begin{equation}
H_d^{(l+1)} =
A_{dw} \left( A_{dw}^\top H_d^{(l)} W_1 + A_{ww} W_2 \right) W_3.
\end{equation}

Finally, the prediction is computed as:
\begin{equation}
\hat{Y} = \mathrm{softmax} \left(
A_{dw} \left( A_{dw}^\top W_1 + A_{ww} W_2 \right) W_3
\right).
\end{equation}

\paragraph{Discussion.}
This formulation removes explicit word-node embeddings while preserving the
original propagation mechanism. The word nodes are implicitly represented
through the composed operators $A_{dw}^\top$ and $A_{ww}$, yielding an
equivalent document-level transformation without maintaining intermediate
word states. As a result, redundant computations are eliminated while
retaining the same decision function for document classification.

\begin{figure}[h]
    \centering
    
    \begin{subfigure}{0.45\textwidth}
        \centering
        \includegraphics[width=\linewidth]{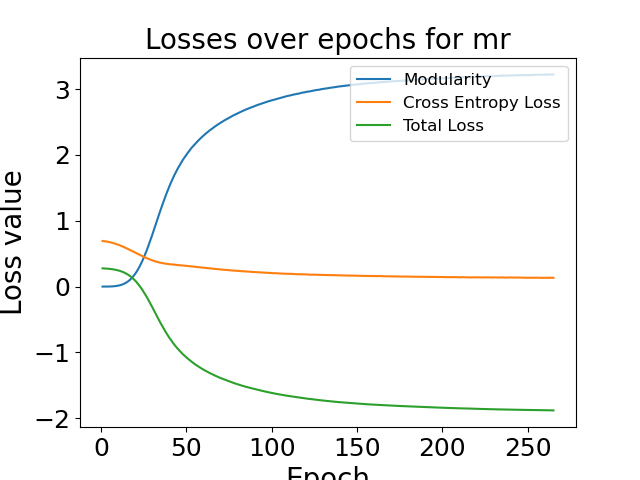}
        \caption{MR}
        \label{fig:sub1}
    \end{subfigure}
    \hfill
    \begin{subfigure}{0.45\textwidth}
        \centering
        \includegraphics[width=\linewidth]{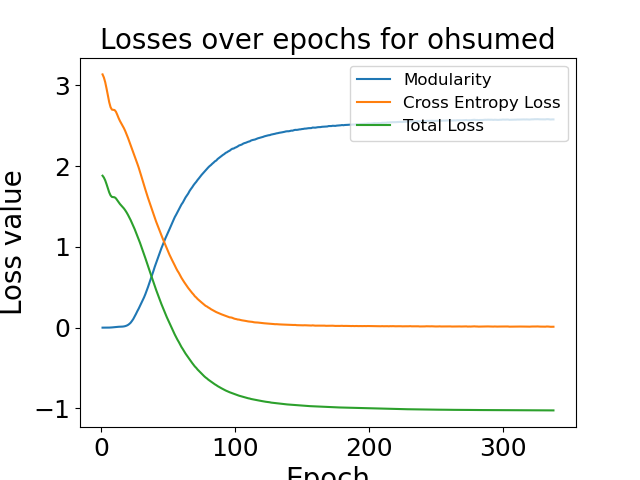}
        \caption{Ohsumed}
        \label{fig:sub2}
    \end{subfigure}

    \vspace{0.4cm}

    \begin{subfigure}{0.45\textwidth}
        \centering
        \includegraphics[width=\linewidth]{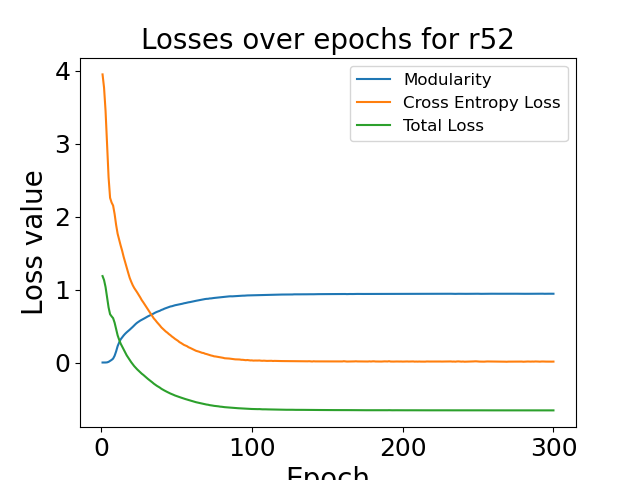}
        \caption{R52}
        \label{fig:sub3}
    \end{subfigure}
    \hfill
    \begin{subfigure}{0.45\textwidth}
        \centering
        \includegraphics[width=\linewidth]{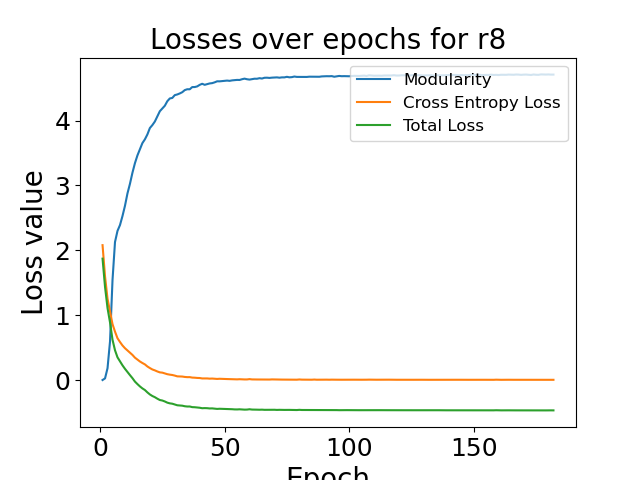}
        \caption{R8}
        \label{fig:sub4}
    \end{subfigure}

    \caption{Training dynamics of ModTGCN on four benchmark datasets (MR, Ohsumed, R52, and R8). Each plot shows the evolution of modularity loss, cross-entropy loss, and total loss across epochs. The consistent decrease in cross-entropy alongside modularity optimization demonstrates stable convergence and highlights the complementary role of global community structure in improving classification performance.}
    \label{fig:losses_interaction}
\end{figure}

\subsection{Robustness of ModTGCN to early pseudo-label noise while training}


Because modularity incorporates predicted labels for unlabeled nodes, we examine sensitivity to early-stage noise using a warm-up strategy that delays the application of the modularity objective for a few initial epochs. As shown in Table \ref{tab:warmup}, the impact of warm-up is consistently negligible across datasets. On MR, R8, and R52, the performance differences remain within $0.3$ points, indicating that early prediction noise does not adversely affect optimization in relatively high-homophily settings. For more structurally complex datasets such as Ohsumed and 20NG, we observe slightly larger gaps ($1.18$ and $1.23$ points, respectively), yet training without warm-up still achieves comparable or better performance. 

Overall, these results suggest that the modularity objective is robust to noisy early predictions and can be applied from the beginning of training without requiring a dedicated warm-up phase.

\begin{table}[t]
\centering
\caption{Effect of modularity warm-up to assess sensitivity to early-stage noise from predicted labels. $\Delta$ denotes the difference between models trained with and without warm-up.}
\label{tab:warmup}
\begin{tabular}{lccc}
\toprule
\textbf{Dataset} & \textbf{With Warm-up} & \textbf{Without Warm-up} & $\boldsymbol{\Delta}$ \\
\midrule
MR      & 88.04 & 88.07 & -0.03 \\
R8      & 98.40 & 98.70 & -0.30 \\
R52     & 96.14 & 96.16 & -0.02 \\
Ohsumed & 76.34 & 77.52 & -1.18 \\
20NG    & 89.91 & 91.14 & -1.23 \\
\bottomrule
\end{tabular}
\end{table}

\begin{table}[h]
\centering
\caption{Comparison of Linear Probing and ModTGCN under pre-trained embeddings (P) across datasets with varying homophily. $\Delta$ Micro-F1 denotes the performance difference (ModTGCN - Linear Probing).}
\label{tab:homophily}
\begin{tabular}{lcccc}
\toprule
\textbf{Dataset} & \textbf{Homophily} & \textbf{Linear Probing (P)} & \textbf{ModTGCN (P)} & $\boldsymbol{\Delta}$ \\
\midrule
MR      & 0.70 & 83.49 & 81.45 & -2.04 \\
R8      & 0.50 & 98.25 & 97.55 & -0.70 \\
R52     & 0.38 & 95.26 & 94.54 & -0.72 \\
Ohsumed & 0.16 & 69.87 & 71.97 & +2.10 \\
20NG    & 0.19 & 79.06 & 90.60 & +11.54 \\
\bottomrule
\end{tabular}
\end{table}
\subsection{ModTGCN: Relationship with homophily}
Compared to embedding-only baselines (Linear Probing (P)), ModTGCN achieves slightly lower performance on simpler, high-homophily datasets (MR, R8, R52), where pretrained representations are already nearly linearly separable, leaving limited room for structural refinement. In contrast, modularity yields substantial gains on structurally complex, low-homophily datasets (Ohsumed $+2.10$, 20NG $+11.54$). This trend indicates that modularity refinement is most beneficial in low-homophily regimes with overlapping semantic boundaries, and less impactful on trivially separable datasets.

ModTGCN shows larger improvements on low-homophily datasets because it incorporates global, degree-corrected structural information through the modularity objective, whereas standard GNNs rely primarily on local neighborhood aggregation. In low-homophily settings, neighboring nodes often belong to different classes, causing local message passing to propagate noisy or conflicting signals and leading to over-smoothing. In contrast, modularity introduces a global coupling mechanism that encourages nodes to align with communities exhibiting higher-than-expected connectivity under a degree-preserving null model. This allows the model to capture long-range, statistically significant structures beyond immediate neighborhoods, while mitigating the influence of hub nodes and noisy edges. As a result, ModTGCN can recover class-consistent communities even when the local graph structure is weak, leading to substantially improved performance in low-homophily regimes.
\end{document}